\documentclass[letterpaper]{article} 
\usepackage{aaai24}  
\usepackage{times}  
\usepackage{helvet}  
\usepackage{courier}  
\usepackage[hyphens]{url}  
\usepackage{graphicx} 
\usepackage{natbib}  
\usepackage{caption} 
\usepackage{algorithm}
\usepackage{algorithmic}
\usepackage{amsmath}
\usepackage{amssymb}
\usepackage{mathtools, nccmath}

\usepackage{newfloat}
\usepackage{listings}
\DeclareCaptionStyle{ruled}{labelfont=normalfont,labelsep=colon,strut=off} 
\floatstyle{ruled}
\newfloat{listing}{tb}{lst}{}
\floatname{listing}{Listing}
\title{Subgoal Discovery \\Using a Free Energy Paradigm and State Aggregations}
\author {
    Amirhossein Mesbah\textsuperscript{\rm 1},
    Reshad Hosseini\textsuperscript{\rm 1},
    Seyed Pooya Shariatpanahi\textsuperscript{\rm 1},
    Majid Nili Ahmadabadi\textsuperscript{\rm 1}
}
\affiliations {
    \textsuperscript{\rm 1}School of ECE, College of Engineering, University of Tehran\\
    amir.mesbah@ut.ac.ir, reshad.hosseini@ut.ac.ir, p.shariatpanahi@ut.ac.ir, mnili@ut.ac.ir
}

\begin{document}

\maketitle

\begin{abstract}
Reinforcement learning (RL) plays a major role in solving complex sequential decision-making tasks. Hierarchical and goal-conditioned RL are promising methods for dealing with two major problems in RL, namely sample inefficiency and difficulties in reward shaping. These methods tackle the mentioned problems by decomposing a task into simpler subtasks and temporally abstracting a task in the action space. One of the key components for task decomposition of these methods is subgoal discovery. We can use the subgoal states to define hierarchies of actions and also use them in decomposing complex tasks. Under the assumption that subgoal states are more unpredictable, we propose a free energy paradigm to discover them. This is achieved by using free energy to select between two spaces, the main space and an aggregation space. The $model \; changes$ from neighboring states to a given state shows the unpredictability of a given state, and therefore it is used in this paper for subgoal discovery. Our empirical results on navigation tasks like grid-world environments show that our proposed method can be applied for subgoal discovery without prior knowledge of the task. Our proposed method is also robust to the stochasticity of environments.
\end{abstract}

\section{Introduction}
Reinforcement learning (RL) ~\cite{sutton2018reinforcement} is widely used in different aspects of our daily life from chatbots ~\cite{christiano2017deep} to autonomous driving ~\cite{kiran2021deep} and chip design~\cite{mirhoseini2020chip}. Classical RL algorithms generally suffer from being time-consuming, sample inefficient, and having difficulties in defining an appropriate reward function. Furthermore, the classical RL algorithms have difficulties in environments with long horizons, delayed rewards, and sparse rewards. Such environments are common in navigation, robotic manipulation, and many other tasks that we are dealing with daily.

Studies like Hierarchical Reinforcement Learning (HRL) ~\cite{make4010009}, Goal-Conditioned Reinforcement Learning (GCRL) ~\cite{liu2022goalconditioned}, and using sub-spaces ~\cite{ghorbani2020reinforcement} are some of the promising efforts that try to solve the aforementioned problems of classical RL algorithms. These works are trying to use a level of abstraction in the action space and the state space or they try to decompose tasks into simpler tasks and use the agent's experience to generalize solving long-horizon tasks.

We can see different examples of action abstraction in our daily lives. For example, instead of thinking about the performance of thousands of pieces of a car, we abstract the sequence of actions taking place into a high-level action like “speed up”. Even more complex tasks like “turning right” can be decomposed into the sequence of lower-level actions “slowing down the speed”, “changing the lane” and “turning the steer to the right”. This process of abstraction can continue recursively. Also, we can decompose tasks like navigation and reaching a special goal state into reaching some defined subgoal states in a row to make the task easier. 

Hierarchical learning which is inspired by learning earning in the human brain~\cite{theves2021learning}, uses abstraction in the action space and takes the power of this property to solve complex tasks faster. Combining hierarchical learning and reinforcement learning has been exploited in three major approaches named option framework ~\cite{sutton1998between}, feudal reinforcement learning ~\cite{dayan1992feudal}, and hierarchical abstract machines ~\cite{parr1997reinforcement}. However, using hierarchical methods is hard in practice and it needs a good knowledge of the task for designing temporally extended actions. In most developed methods, especially the options framework, besides learning a hierarchy of policies the most challenging problem is option discovery, to find subgoal or bottleneck states to create options.

\begin{figure*}[t]
\centering
\includegraphics[width=0.95\textwidth]{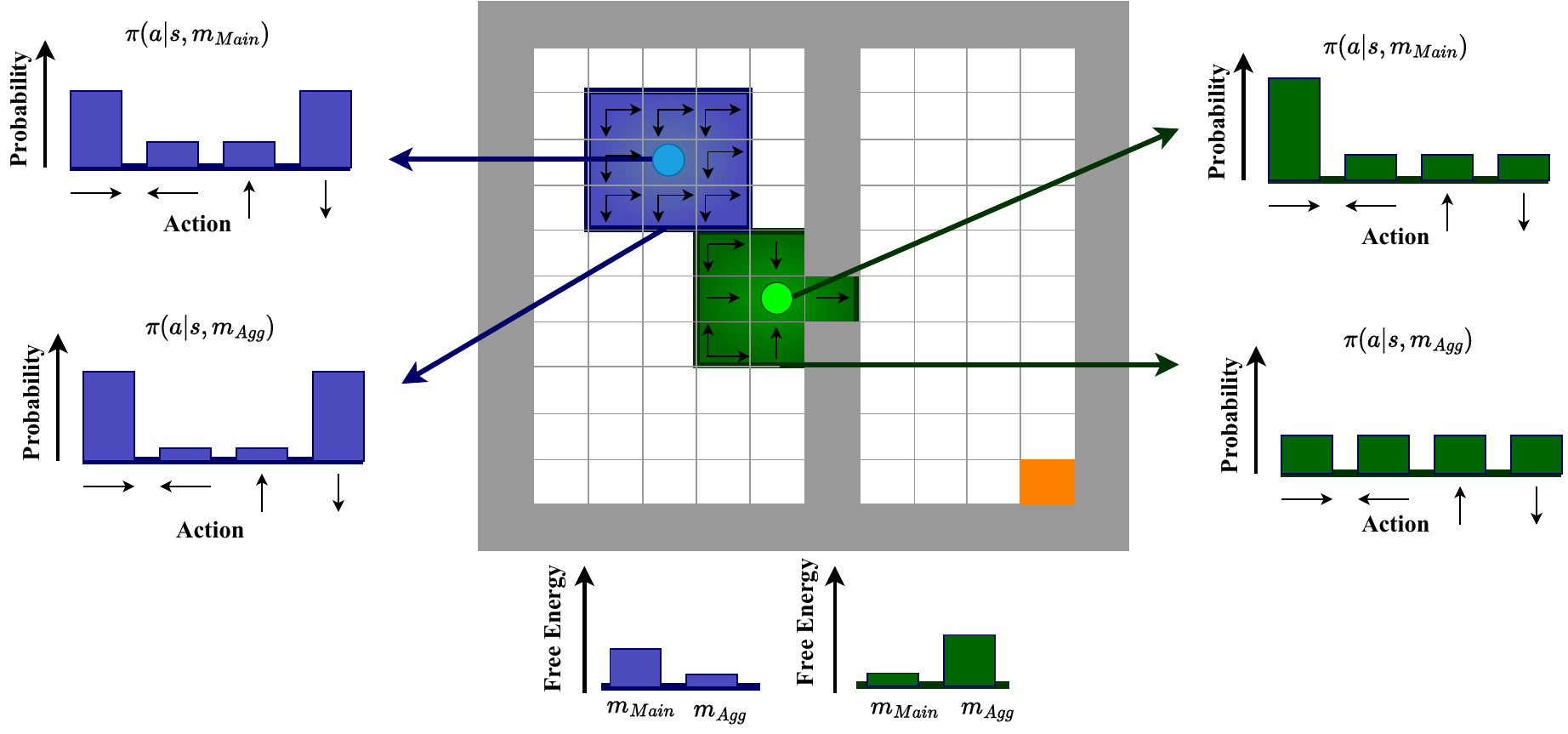} 
\caption{This figure illustrates a two-room environment with a doorway connecting them and a goal in the bottom right (orange). An agent starting from the top left can move in four directions. Each state in the main space corresponds to a 3x3 block of states in the aggregation space. The agent chooses between main space ($\pi(a|s,m_{Main})$) and aggregation space ($\pi(a|s,m_{Agg})$) policies based on their Free Energy (uncertainty measure) - selecting the space with a lower free energy. Near the bottleneck (doorway) states, the main space is preferred due to its lower free energy, while the aggregation space is chosen in states distant from doorways.}
\label{fig1}
\end{figure*}

A similar approach to HRL is GCRL. In this setting an agent can have a high-level controller for finding appropriate subgoals in the environment. After detecting subgoals, a low-level policy can learn the sequence of subgoals and simultaneously decompose a complex task into simpler ones ~\cite{nachum2018dataefficient, gupta2019, chanesane2021goalconditioned}. Therefore, having a method to detect subgoal states in the environment can help with option discovery in HRL and for learning implicit or explicit subgoal-based policies in goal-conditioned settings.

As mentioned above, the other level of abstraction in an environment is state abstraction which leads to approaches like using sub-spaces or state aggregation and grouping~\cite{daee2014reward, hashemzadeh2018exploiting, hashemzadeh2020clustering, li2023neural}. Using sub-spaces and state aggregations can provide us with more samples from the environments and lead to a speed-up in exploration and learning in the initial episodes. We can lower the effect of sample inefficiency using sub-spaces which are many-to-one mapping from the main environment. In these kinds of tasks, selecting which states to aggregate and which sub-spaces to choose for learning in each step of episodes are the challenging questions. Furthermore, aggregating different states in state space can encounter the problem of perceptual aliasing (PA). This problem happens when the aggregated states have totally different policies and this causes the samples in aggregation spaces to become useless.

In this paper, we are interested in dealing with the question “How can we identify bottleneck states faster using state aggregations during an artificial RL agent’s lifetime?”. Using the state aggregations to detect bottlenecks or subgoals faster is the main motivation of our work. As illustrated in Figure~\ref{fig1}, the aggregation of states in a bottleneck state causes an increase in the uncertainty of policy in the aggregation space and makes its policy close to a random policy. This is because of the PA problem and the difference in the policy of aggregated states. However, in states far from the bottleneck, we can use aggregation to improve the samples in the direction of the optimal policy. In bottleneck states, our considered method of aggregation does not help improve learning, and this transformation is not smooth in bottleneck states. 

We present an algorithm to identify subgoal states by capturing this uncertainty using an information-theoretic concept called free energy. Our proposed algorithm can identify the bottleneck states faster than other methods which are mostly based on creating a graph of the environment. Also, there is no need to define the number of subgoals compared to methods with differentiable termination functions for the termination condition of options. 

In summary, this paper provides the following contributions: (a) We propose a new bottleneck detection algorithm that identifies bottleneck states after living some episodes in the environment. (b) We demonstrate that our proposed method is empirically robust to the stochasticity of the environment and our method can identify subgoal states in the environment with up to 50\% of stochasticity. (c) Our idea of model changes applies to both discrete and continuous state spaces considering some modifications. In the following sections, we review the literature and state our assumptions and preliminary findings. We then present our method and experimental results. Lastly, we discuss future research directions and conclusions.

\begin{figure*}[t]
\centering
\includegraphics[width=0.7\textwidth]{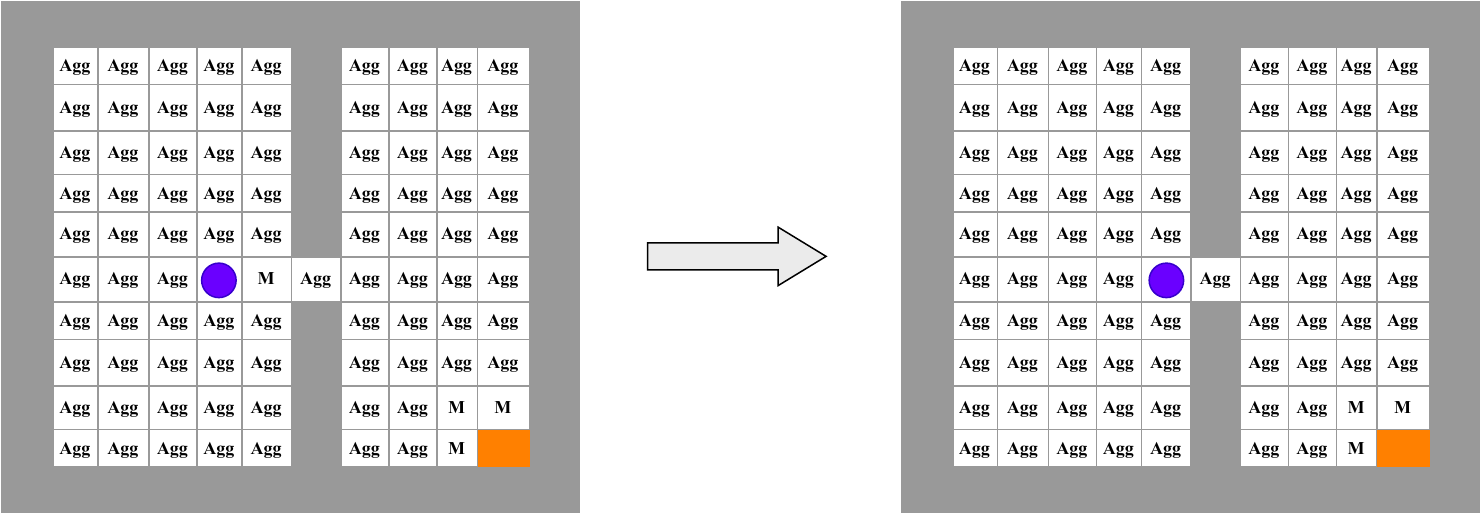} 
\caption{This figure demonstrates the agent's state space selection in a two-room environment. The blue circle represents the agent's current position, moving toward the goal (orange). States are assigned either $Main (M)$ or $Aggregation (Agg)$ space based on the minimum free energy. Near and at the doorway, the agent switches from Aggregation to Main space due to higher uncertainty in aggregated states at bottlenecks, illustrating the $model\; changes$ during navigation.}
\label{fig2}
\end{figure*}

\section{Related Work}
The studies on bottleneck or subgoal discovery can be categorized into two groups, GCRL and option discovery in HRL. In GCRL methods, a high-level controller tries to identify subgoals. The detected subgoals are used as the target of low-level policies \cite{chanesane2021goalconditioned, nachum2018dataefficient, nair2019hierarchical} for learning decomposed tasks. These methods are implemented on the goal-conditioned Markov Decision Processes (MDPs) where we have the information about the goal state and our agent is trying to learn an optimal policy to reach the desired goal. In contrast to the assumption considered in these methods, we consider the vanilla MDP setting without providing any additional information on the goal state for our agent. 

The other group of ideas for the subgoal detection problem are referred to as the option discovery problem in the HRL setting. This is one of the challenging problems in this framework alongside learning hierarchical policies ~\cite{make4010009}. It is important to note that option discovery is referred to as skill discovery in some cases. However, the definition and intuition of skill are different from options in the RL literature ~\cite{eysenbach2018diversity}.

Methods for creating new options must determine when to create an option, how to define its termination condition (skill discovery), how to expand its initiation set, and how to learn its policy. Consequently, option discovery methods can be categorized into three main groups: implicit option learning, gradient-based option discovery, and option learning approaches based on finding bottleneck states or subgoals.

Implicit option learning methods try to implicitly learn the options by augmenting the proposed semi-MDP into a new MDP by adding options to the state space, and actions of selecting options to the action space. 
In addition to this, a binary random variable $\beta$ corresponding to each option is added to action space which is a termination condition with a value of True or False. This illustrates an option has to be terminated when its corresponding binary variable is True ~\cite{levy2011unified}. In ~\cite{daniel2016probabilistic}, they used probabilistic inference methods to infer the termination of an option in the augmented MDP setting. In this framework, the number of options has to be determined before the agent starts the learning process which may need domain knowledge in some environments, and it is hard to apply these methods in continuous domains.

Gradient-based option discovery methods try to define a policy gradient theorem in the options framework considering the option-state value and termination function~\cite{bacon_option-critic_2016}. In ~\cite{smith2018inference}, options are considered as latent variables and they are trained through policy gradient. ~\citet{harb_when_2017} used a deliberation cost for learning options in an option-critic framework. Using information-theoretic objectives to learn a diverse set of options was proposed in ~\cite{kamat_diversity-enriched_2020}. In ~\cite{levy2017learning, fox2017multi, riemer2018learning, zhang_dac_2019, wan_toward_2022}, the authors increased the levels of hierarchy to more than two by proposing new option-critic architectures. Furthermore, ~\citet{khetarpal2020options} proposed to learn the initial condition of an option in addition to each option’s policy and termination condition. In general, gradient-based option discovery methods have shown very good performance, especially in continuous-state space environments. But, they have some drawbacks like being slow in finding options and requiring to fix the number of options before learning.

Option learning methods consider bottleneck states to be more informative and use them to design options automatically. These methods use information like trajectories and acquired rewards to find bottlenecks and design options. These methods usually construct a graph of the environment at the initial steps. They then use graph theoretic methods to find bottlenecks, the methods like min/max flow (Q-cut)~\cite{menache2002q}, minimizing the graph's cover time~\cite{jinnai2019discovering, jinnai2019exploration,jinnai2019skill, jinnai2019finding}, graph clustering~\cite{mannor2004dynamic}, local graph partitioning (L-cut)~\cite{csimcsek2005identifying}, betweenness~\cite{csimcsek2008skill}, strongly connected components~\cite{kazemitabar2009using}, min degree and max distance~\cite{zhu2022mdmd}, and graph Laplacian, by using proto-value functions~\cite{machado2017laplacian, mendoncca2019laplacian}. Apart from graph theoretic methods, there are other methods that use heuristics and techniques for subgoal detection. ~\cite{mcgovern2001automatic} is one of the first works on option learning that incorporates the diverse density heuristic by collecting successful and unsuccessful trajectories and defining states with high frequency that always appear in successful trajectories. Some notable works for option learning use different approaches for option learning, like skill chaining to discover options in continuous state spaces~\cite{konidaris2009skill}, action restriction (states with unique action direction as subgoals)~\cite{ding2014autonomic}, calculating occurrence probability~\cite{pateria2021end} and access states with relative novelty for each state~\cite{csimcsek2004using}.

Graph theoretic and trajectory-based algorithms have some shortcomings in finding bottlenecks. Constructing graphs on the environment using trajectory can be time-consuming and inefficient. Also creating a graph may have problems in the environment with a high rate of stochasticity and this may cause to construction of an inaccurate graph of the environment. ~\citet{rafati2019efficient} tried to overcome these shortcomings by finding subgoals using unsupervised  
anomaly detection with the k-means algorithm on experience memory. \citet{ramesh2019successor} used successor representations to identify successor options. ~\citet{manoharan2021option} used an autoencoder to detect the subgoal of options.

Using sub-spaces has not been studied for subgoal detection, but we shortly review them here, because our method is based on the idea of using sub-spaces. In ~\cite{hashemzadeh2018exploiting}, the role of incorporating sub-spaces in the learning process, especially in the initial episodes was studied. Subspaces have the PA problem, to deal with this problem a clustering approach is used in ~\cite{hashemzadeh2020clustering}. A powerful and general framework on how to integrate the decision of the subspaces and the main space was proposed in~\cite{ghorbani2020reinforcement} by incorporating a free energy paradigm. In this work, we use the free energy in our proposed framework for bottleneck discovery. Furthermore, our way of defining the subspaces is different from these previous works.

\begin{figure*}[t]
\includegraphics[width=1\textwidth]{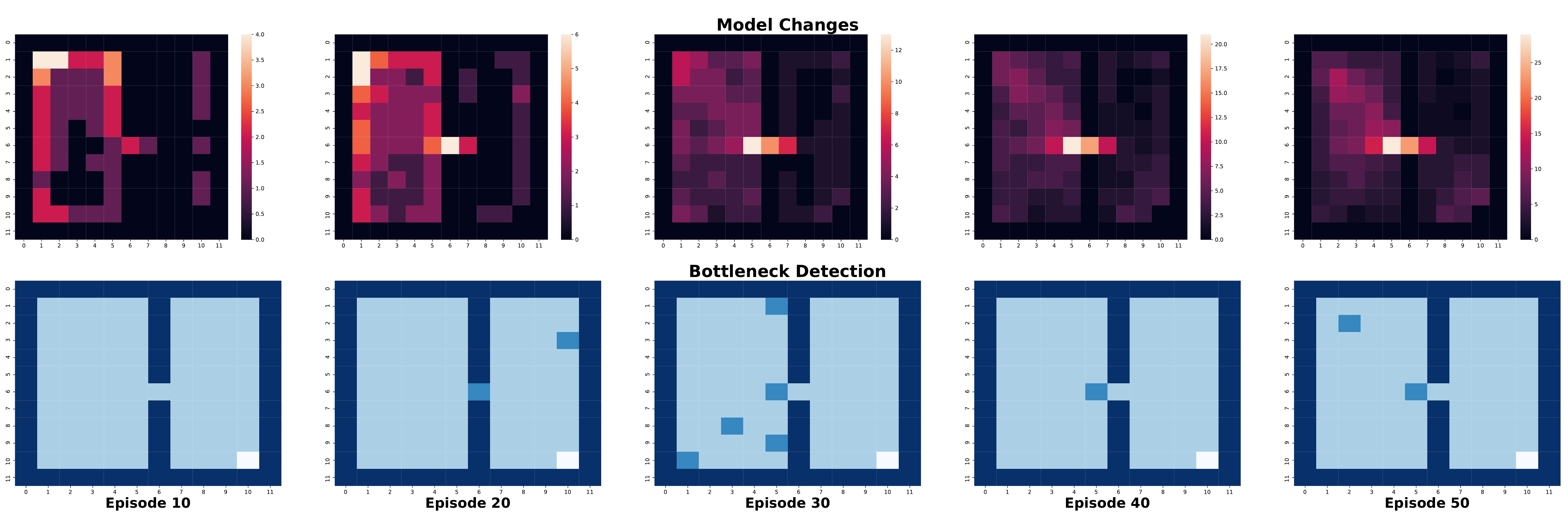} 
\caption{This figure tracks the incremental process of model changes from episodes 10 to 50 in a two-room environment, where the agent starts from the upper left corner and aims to reach a white goal state in the bottom right. The top row shows the number of model changes in transitioning from one state to another, while the bottom row shows identified bottlenecks considering the model changes. By episode 40, the agent accurately identifies the doorway as a bottleneck, and by episode 50, it also recognizes key states along the optimal path from the top-left start to the bottom-right goal.}
\label{fig3}
\end{figure*}

\section{Background}

\subsection{Reinforcement Learning}
RL is a framework for solving sequential decision-making tasks in a trial-error manner. Environments in this framework are modeled with a MDP defined by a tuple $<S, A, R, P, \gamma>$ where $S$ is the state space, $A$ is the action space, and $R$ is the expected reward. $P$ specifies the dynamics of the environment and $\gamma$ is the discount factor. Assume in a MDP at time step $t$, the agent in state $s_t \in S$ commits action $a_t \in A$ on the environment, the next state of the agent in the environment $s_{t+1} \in S$ is determined by the transition probability $P_a(s, s^{'})=Pr(s_{t+1}=s^{'}|s_t=s, a_t=a)$, and the agent receives the instant reward $r_{t+1}$, whose its expectation is equal to $R_a(s, s^{'})$. In most of RL problems, the agent does not know the model of the environment, and it needs to interact with the environment to learn a policy $\pi: S\times A\rightarrow[0, 1]$ during its lifetime in the environment and it is trying to find an estimate to an optimal policy $\pi^*$. An optimal policy in each state is a policy that maximizes the cumulative expected reward given by
\begin{equation*}
\begin{aligned}
    G_t =  \sum_{k=t}^{\infty} \gamma^k r_{t+k+1}.
\end{aligned}
\end{equation*}

In this work, we consider model-free algorithms where the agent does not directly learn the model of the environment, and instead, it calculates an estimate to the action-state value function $Q_{t}(s, a)$:
\begin{equation*}
\begin{aligned}
     Q_{t}(s, a) &=  \mathbb{E}_{\pi}[G_t| s, a] \\
        &= \sum_{s^{'}, r}P_{a}(s, s^{'})(R_a(s, s^{'}) + \gamma \sum_{a^{'}} \pi(a^{'}|s^{'})Q(s^{'}, a^{'})),
\end{aligned}
\end{equation*}
using algorithms like single step SARSA ~\cite{sutton2018reinforcement} with the following update rule given that in time step $t$ we are in state $s$ and we take action $a$ using our policy:
\begin{equation*}
\begin{aligned}
    Q_{t+1}(s, a) &= Q_{t}(s, a) + \lambda(r_{t+1} + \gamma Q_{t}(s^{'}, a^{'}) - Q_{t}(s, a)),
\end{aligned}
\end{equation*} 
where $\lambda$ is the learning rate, $\gamma$ is the discount factor of the environment, $s^{’}$ is the next state upon taking the action, $r_{t+1}$ is the instant reward, and $a^{’}$ is the next action that is chosen based on our policy in the next state. SARSA is an on-policy algorithm that chooses $a^{'}$ with respect to its policy in the next state $s^{'}$.

\subsection{Free Energy}
The free energy paradigm is related to the laws of thermodynamics that explain why energy flows in certain directions. We can use this paradigm for describing a system or making predictions. The free energy concept that we use in decision-making and neuroscience was introduced by~\cite{friston2006free} and then developed by ~\cite{ortega2015informationtheoretic} for rationally bounded decision-making. From ~\citet{friston2006free} perspective, the human brain is trying to minimize a variable called free energy which is the same as minimizing a surprise function or maximizing evidence of the model of the environment. To achieve this goal, we need to have a good model of the environment.

If we model the observations of the environment using a random variable $O$, and the hidden state of the environment with the random variable $S$ (we do not have access to the real state of the environment), our brain constructs a generative model defined as $P(O, S)$. We can calculate surprise with $-\log(P(O))$. As the probability of our observation $O$ becomes high (near 1), the surprise of this observation becomes less. Considering a dummy distribution $Q(s)$ we can have:
\begin{equation*}
\begin{aligned}
    -\log P(O)=-\log \sum_{s \in S} P(O, s) =-\log \sum_{s \in S} Q(s) \frac{P(O, s)}{Q(s)}.
\end{aligned}
\end{equation*}
The considered surprise function is convex, so we can use Jensens’s inequality: 
\begin{equation*}
\begin{aligned}
    f(w x+(1-w) y) \leq w f(x)+(1-w) f(y),
\end{aligned}
\end{equation*}
to calculate an approximate upper bound for the surprise function.
So we have:
\begin{equation*}
\begin{aligned}
    -\log \sum_{s \in S} Q(s) \frac{P(O, s)}{Q(s)} &\leq-\sum_{s \in S} Q(s) \log \frac{P(O, s)}{Q(s)} \quad \\
    &=\sum_{s \in S} Q(s) \log \frac{Q(s)}{P(O, s)}.
\end{aligned}
\end{equation*}
This upper bound for the surprise function is called variational free energy $F$, which can be expanded as below:
\begin{equation}\label{eq2}
\begin{aligned}
    F  & =\sum_s Q(s) \log \frac{Q(s)}{P(s) P(O \mid s)} \\ & =K L(Q(s) \| P(s))-\sum_s Q(s) \log P(O \mid s),
\end{aligned}
\end{equation}
where $KL$ measures the KL divergence between Q and P.
The first term of \eqref{eq2}, named complexity, shows how much the approximation of the posterior (any arbitrary dummy distribution $Q(s)$)  deviates from the prior $P(s)$. The second term, named accuracy, shows how likely are the states $s$ given a specific outcome $O$. We will use this concept to determine which space is predictable and to detect an increase in uncertainty.

The free energy in decision-making when the agent is rationally bounded was studied by ~\citet{ortega2015informationtheoretic}. They derived a free energy formulation for single-step and sequential decision-making problems. In this formulation, the authors combined a linear and a non-linear cost function for calculating the free energy. The linear cost function is related to the expected utility and the non-linear cost function is equal to the KL-divergence of prior and posterior choice probabilities as a measure of the information cost. The expected utility term is the same as the accuracy term in \eqref{eq2}. The advantage of this formulation was that it has a closed-form optimal solution.

\subsection{Aggregating different perspectives in RL using Free Energy}
Aggregating different states, i.e. subspaces, has shown a notable enhancement in the performance of agents. In addition to speeding up learning, especially at the initial steps, using subspaces can lead to enhancing sample efficiency as well as lowering regret \cite{ghorbani2020reinforcement, hashemzadeh2018exploiting, hashemzadeh2020clustering}. However, defining subspaces opens up different research questions like "which states to aggregate as a subspace", "How to cope with PA problem in subspaces" and "which subspaces to choose for learning at each step".  

In the most recent work, \citet{ghorbani2020reinforcement} introduced a free energy based approach for choosing the best defined subspace for learning at each step. They have supposed a utility function based on the Thompson sampling policy \cite{russo2018tutorial}, as the negative informational surprise:
\begin{equation}\label{eq3}
\begin{aligned}
    U(a, s, m) = \log\pi_{T S}(a \mid s, m).
\end{aligned}
\end{equation} 
This utility gives information about the optimality of action $a$ at state $s$ and subspace $m$.

To utilize subspaces for learning and avoiding the PA problem, the following constraint is used to ensure the utility of the main space is close to the utility of subspaces for any target policy $\pi$:
\begin{equation}\label{eq4}
\begin{aligned}
    \mathbb{E}_{\pi(a|s,m)}[U(a,s,m)] - \mathbb{E}_{\pi(a|s,m)}[U(a,s,m_{Main})] < K_1,
\end{aligned}
\end{equation} 
where $m$ is any considered subspace and $m_{Main}$ is the main space. Also, to lower the effect of inaccurate uncertainty estimation, another constraint is used to limit the policy by an arbitrary behavioral policy $\pi_{B}$:
\begin{equation}\label{eq26}
\begin{aligned}
    D_{KL}(\pi(a|s,m)||\pi_B(a|s,m)) < K_2.
\end{aligned}
\end{equation}
Considering the defined utility function and mentioned constraints, the problem of learning the optimal policy utilizing different subspaces changes into the following optimization problem which is a free energy minimization, similar to \eqref{eq2}:
\begin{equation} \label{eq5}
\begin{aligned}
    \pi^{*}(a| s, m) = \arg \min_{\pi(a|,s, m)}  F(s, m, \pi(a | s, m)),
\end{aligned}
\end{equation}
where the free energy for any target policy $\pi(a| s, m)$ and for each state $s$ and space $m$ is given by
\begin{equation} \label{eq6}
\begin{aligned}
    &F(s, m, \pi(a| s, m)) = \mathbb{E}_{\pi(a|s, m)} [ \frac{1}{\alpha} \log \frac{\pi(a|s, m)}{\pi_{B}(a|s, m)} \\
    &+ \frac{1}{\beta} \log \frac{\pi(a|s, m)}{\pi_{TS}(a|s, m_{Main})}  - \log \pi_{TS}(a|s, m)].
\end{aligned}
\end{equation}
There is a closed-form solution for $\pi^{*}$ in \eqref{eq5}, that is given by
\begin{equation}\label{eq7}
\begin{aligned}
     &\pi^{*}(a|s, m) = \frac{1}{z(s, m)} \pi_{B}(a|s, m) e^{\alpha {\hat{U}(a, s, m)}}, \\
     &z(s, m) = \sum\limits_{a} \pi_{B}(a|s, m) e^{\alpha {\hat{U}(a, s, m)}},
\end{aligned}
\end{equation} 
wherein
\begin{equation}
\begin{aligned}\label{eq8}
   \medmath{\hat{U}(a, s, m) = U(a, s, m) - \frac{1}{\beta}(U(a, s, m) - U(a, s, m_{Main}))}.
\end{aligned}
\end{equation}

\section{Subgoal discovery using State Aggregations and Free Energy paradigm}

In this paper, we consider the concept of bottleneck to define options in the environment. States like doorways in multi-room environments can be seen as subgoal states, so identifying such states can help us to decompose the task of navigation from one room to the other or it can be beneficial for autonomous option discovery in the options framework.

To detect such states, we assume our agent lives in an environment with a defined state space and we call this space $Main \; Space$:
\begin{equation*}
\begin{aligned}
    m_{Main} : \phi_{Main} (s) = s , s \in S,
\end{aligned}
\end{equation*}
where $\phi_{Main}$ is an identity function. We update the Q-values of the main space with the update rule of our learning algorithm, SARSA:
\begin{equation}
\begin{aligned}\label{eq9}
    & Q_{Main}(s, a) =  Q_{Main}(s, a) \\
    & + \lambda (r + \gamma Q_{Main}(s^{’}, a^{’}) - Q_{Main}(s, a)),
\end{aligned}
\end{equation}
where $\lambda$ is the learning rate, and $r$ is the instant reward that the agent gets from the environment. Also, $(s^{'}, a^{'})$ indicates the next state-action pair.

In addition to the main space, we assume our agent has access to its physical neighbor states and their action-state values. So we can define an $Aggregation \;  Space$ ($m_{Agg}$):
\begin{equation*}
\begin{aligned}
    m_{Agg}: \phi_{Agg} (s) =\{s’ | s’ \in S \; \& \; d(s, s’) < L\} , s \in S,
\end{aligned}
\end{equation*}
where $d$ is a distance metric like Euclidean or Manhattan distance. Also, $L$ is the maximum distance of the neighborhood of the current state.
We don’t have any learning in the aggregation space and the Q-values of this space are calculated using a weighted average on aggregated states’ Q-value in the main space:
\begin{equation}
\hspace{-0.cm}
\begin{aligned}\label{eq28}
    Q_{Agg}(s, a) &= \frac{1}{{ \sum\limits_{s^{'} \in \phi_{Agg} (s)}} n_{Main}(s^{'}, a)} \times \\
    &\sum_{s^{'} \in \phi_{Agg} (s)} n_{Main}(s^{'}, a) \times  Q_{Main}(s^{'}, a).
\end{aligned}
\end{equation}
In this equation, $n_{Main}(s, a)$ gives the frequency of samples of action $a$ in state $s$ in the main space ($m_{Main}$).

Following \cite{ghorbani2020reinforcement}, the best space between the main and the aggregation space is the one that minimizes the free energy given by~\eqref{eq5}, mathematically speaking:

\begin{equation}\label{eq29}
\begin{aligned}
    m^{*}(s) = \arg \min_{m}  F(s, m, \pi^{*}(a | s, m)),
\end{aligned}
\end{equation}
where $\pi^{*}$ can be calculated by \eqref{eq7} for each space.

Thompson sampling policy is equal to
\begin{equation}
\begin{aligned}\label{eq10}
    \pi_{T S}(a_i \mid s, m)=P(&\bigcap\limits_{j \neq i}\{Q_{m}(s, a_i)\\
    &>Q_{m}(s, a_j)\}),
\end{aligned}
\end{equation}
where $Q_{m}(s, a)$ is the belief distribution for the value of the action $a$ in state $s$ and space $m$. Calculating the exact belief distribution for each state-action value in each space is computationally complex and hard to achieve. Thus, we can apply approximation methods by calculating an upper and lower bound for these distributions ~\cite{AUDIBERT20091876}, by considering a confidence interval of $1-\nu$:
\begin{equation}
\begin{aligned} \label{eq11}
   \medmath{ P(\hat{Q}_{m}(s, a)-\mu<Q_{m}(s, a)<\hat{Q}_{m}(s, a)+\mu) \geq 1-\nu},
\end{aligned}
\end{equation}
where $\mu$ is given by
\begin{equation}\label{eq12}
\begin{aligned} 
    \medmath{\mu= \overline{std}(s, a, m)\sqrt{\frac{2\log \frac{3}{\nu}}{n_{m}(s, a)}} + \frac{3\log 
    \frac{3}{\nu}}{n_{m}(s, a)}},
\end{aligned}
\end{equation}
and std is computed by
\begin{equation}
\begin{aligned}\label{eq13}
\medmath{\overline{s t d}(s, a, m)=\sqrt{\frac{n_{m}(s, a) \sum_t \tilde{Q}_{t, m}(s, a)^2-\left(\sum_t \tilde{Q}_{t, m}(s, a)\right)^2}{n_{m}(s, a) \times(n_{m}(s, a)-1)}}}.
\end{aligned}
\end{equation}
In this equation, $\tilde{Q}_{t, m}$ is the value of action $a$ in state $s$ and space $m$ at the step $t$ of the agent's lifetime.

We now have the ingredients to implement our idea of bottleneck discovery as explained in the introduction (see Figs \ref{fig1},\ref{fig2}). We defined model changes as a measure of the irregularity of a state. If by entering a state from its neighboring states, there is a change between the best spaces, we count up the value of the model change of that particular state, that is
\begin{equation} \label{eq14}
        MC(s)=
        \left\{ \begin{array}{ll}
            MC(s) + 1 & m^{*}(s_{t-1})\neq m^{*}(s_{t}), \\
            MC(s) &\text{otherwise}.
        \end{array} \right.
    \end{equation}
In \eqref{eq14}, $m^{*}(s_{t})$ is the free energy model of the environment for the current state of the agent and $m^{*}(s_{t-1})$ is the free energy model of the previous state.

\begin{figure*}[t]
\includegraphics[width=1\textwidth]{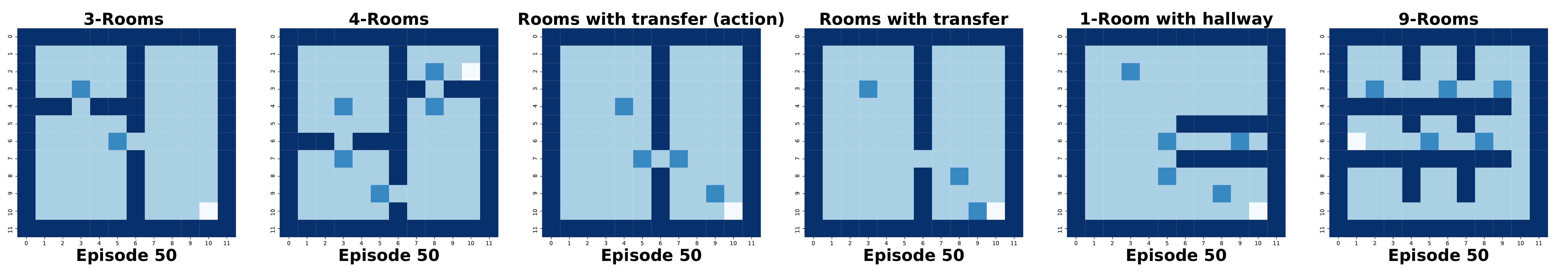} 
\caption{This figure shows detected bottlenecks (in light blue) across six different environments at episode 50, using a state aggregation distance of L=2 for aggregation. In each environment, the agent starts from the upper left corner and aims to reach a white goal state in the bottom right. In the 1-room with hallway environment, bottlenecks appear along the hallway and near the goal due to early exploration patterns and low model changes throughout this environment. While these hallway bottlenecks could be useful for defining "leaving hallway" options, increasing L could prevent their detection in case they are undesired.}
\label{fig5}
\end{figure*}

\begin{figure}[H]
\includegraphics[width=0.95\columnwidth]{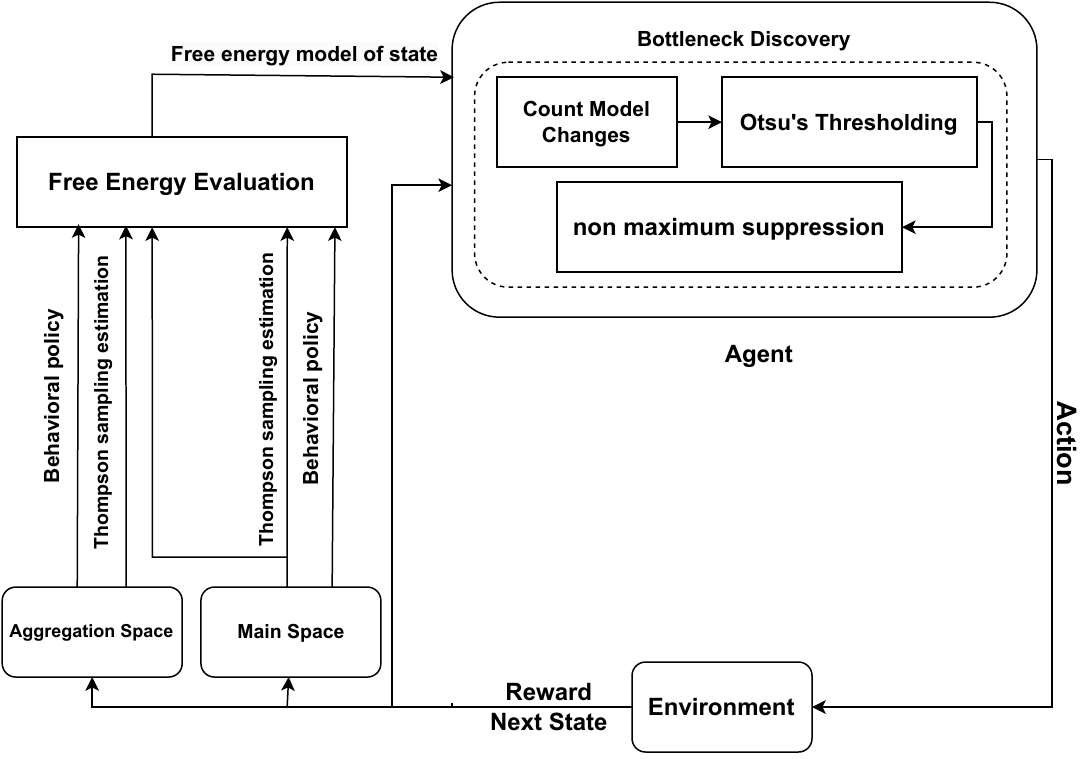} 
\caption{This diagram illustrates the proposed method. The system evaluates free energy in both aggregation and main spaces using an approximate estimation of Thompson sampling and behavioral policy. The bottleneck discovery module tracks model changes between states, applies Otsu's thresholding, and uses non-maximum suppression to identify bottleneck states. The agent interacts with the environment, receiving rewards and next states while updating its state space model based on free energy evaluation.}
\label{fig4}
\end{figure}

Figure \ref{fig4} shows the flow of our proposed method. In each step of the episode, we first estimate the Thompson sampling policy by approximating the intervals of state-action values for each space, which is computed by \eqref{eq10}. After this step, we calculate the free energy of each space to decide which space has minimum free energy, computed by \eqref{eq5}. If the free energy model of the current state is different than the previous state, we will count for model changes in the current state, as expressed by \eqref{eq14}. For the aim of bottleneck detection, we apply Otsu's thresholding ~\cite{otsu1979threshold} to determine the states with a higher count of model changes considering model changes as a matrix of all states. This thresholding algorithm is variance-aware and it clusters states into two groups of zeros and ones. The output of this thresholding method is a matrix with values of zero or one for each state. By piecewise multiplying the output of Otsu's thresholding algorithm and model changes matrices, and finally applying a non-maximum suppression on the result matrix, we can identify bottleneck states.

Algorithm~\ref{alg1:algorithm} specifies the pseudocode of our proposed method. Also, the computational complexity of calculating free energy models for each space is $O(|S||A|^2)$, where $|S|$ denotes the cardinality of the state space and $|A|$ is the cardinality of the action space. 

The idea of model changes can be used in environments with continuous state space with some modifications. Considering an infinite number of states in these kinds of environments, for calculating Q-values in aggregation space, we can sample from the neighborhood of the current state by considering a maximum distance and a distance measure. Also, to calculate the Q-value in the aggregation space, we need sample counts of each action in each neighbor state. We can use the experience of the agent which is saved in the replay buffer for this purpose.

\begin{algorithm}[H]
\caption{Our Proposed Algorithm}
\label{alg1:algorithm}
\begin{algorithmic}[1] 
\FOR{Each Episode}
\WHILE{$done\neq True$} 
\STATE next state, reward, done = environment.step(action)
\STATE calculate free energy model of state, using \eqref{eq5}
\IF {$m^{*}(state) \neq m^{*}(previous \; state)$}
\STATE Count for model change in state, using \eqref{eq14}
\ENDIF
\ENDWHILE
\STATE Apply Otsu's thresholding on model changes matrices
\STATE Non maximum suppression on $model \; changes \times Otsu's \; thresholding \; output$
\ENDFOR
\STATE \textbf{return} bottleneck states
\end{algorithmic}
\end{algorithm}

\begin{algorithm}[H]
\caption{Model Changes in Continuous State Space}
\label{alg2:algorithm}
\begin{algorithmic}[1] 
\FOR{Each Episode}
\WHILE{$done\neq True$} 
\STATE next state, reward, done = environment.step(action)
\STATE Sample from neighbor states with a distance of $L$
\STATE Calculate Q-values for aggregation space using replay buffer
\STATE Calculate Thompson sampling by considering the frequency of actions in the replay buffer
\STATE Calculate free energy model of state, using \eqref{eq5}
\IF {$m^{*}(state) \neq m^{*}(previous \; state)$}
\STATE Count for model change in state, using \eqref{eq14}
\ENDIF
\ENDWHILE
\ENDFOR
\STATE \textbf{return} Model changes for each state
\end{algorithmic}
\end{algorithm}

In continuous state spaces, it is possible to calculate an approximation of a belief distribution by applying dropout before each weighting layer of the Q-network. Thus, we can calculate the Thompson sampling policy using \eqref{eq15}, where $N$ is the number of networks estimating Q-values for action $a_i$ in state $s$ and $n_s$ is the number of times that action $a_i$ has been selected in state $s$ ~\cite{ghorbani2020reinforcement, pmlr-v48-gal16}:

\begin{equation}\label{eq15}
\begin{aligned}
     \pi_{TS}(a_{i}|s) = \frac{n_{s}(a_{i},s)}{N}.
\end{aligned}
\end{equation}
Algorithm~\ref{alg2:algorithm} provides the modification of our developed algorithm for model changes in the continuous domain.

\begin{figure}[H]
\hspace{-0.1cm}
\includegraphics[width=1\columnwidth]{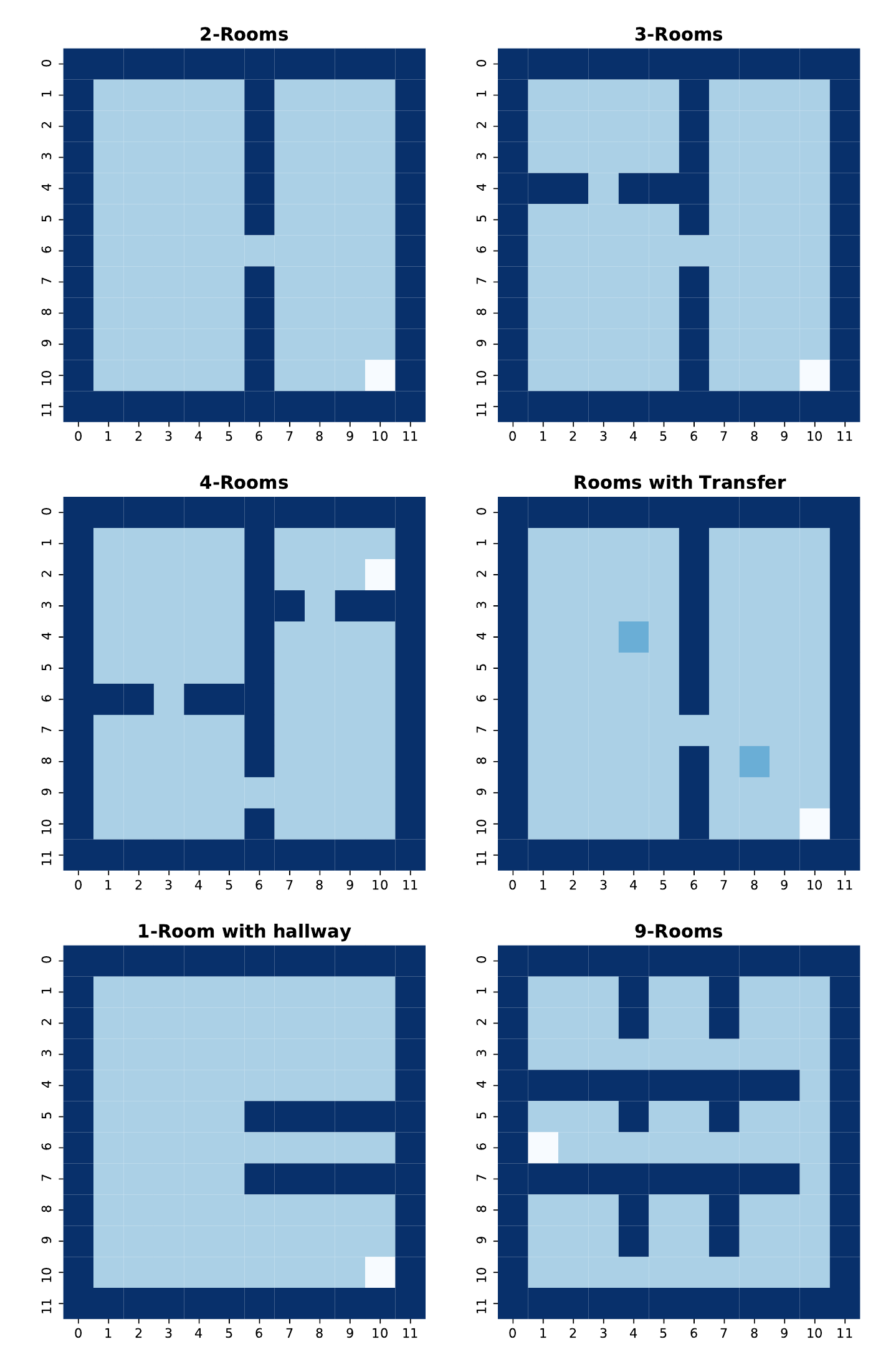} 
\caption{This figure shows six different grid-world environments used in the experiments: 2-rooms, 3-rooms, 4-rooms, rooms with a transfer state (in two variants), 1-room with a hallway, and 9-rooms. In each environment, the white cell represents the goal state. The rooms with transfer state environment has two versions: one with a special transfer action that is activated in state (4, 4), and another where stepping into state (4,4) automatically teleports the agent to state (8,8). }
\label{fig6}
\end{figure}

\section{Experimental Results}

\begin{figure}[H]
\hspace{-0.1cm}
\includegraphics[width=1\columnwidth]{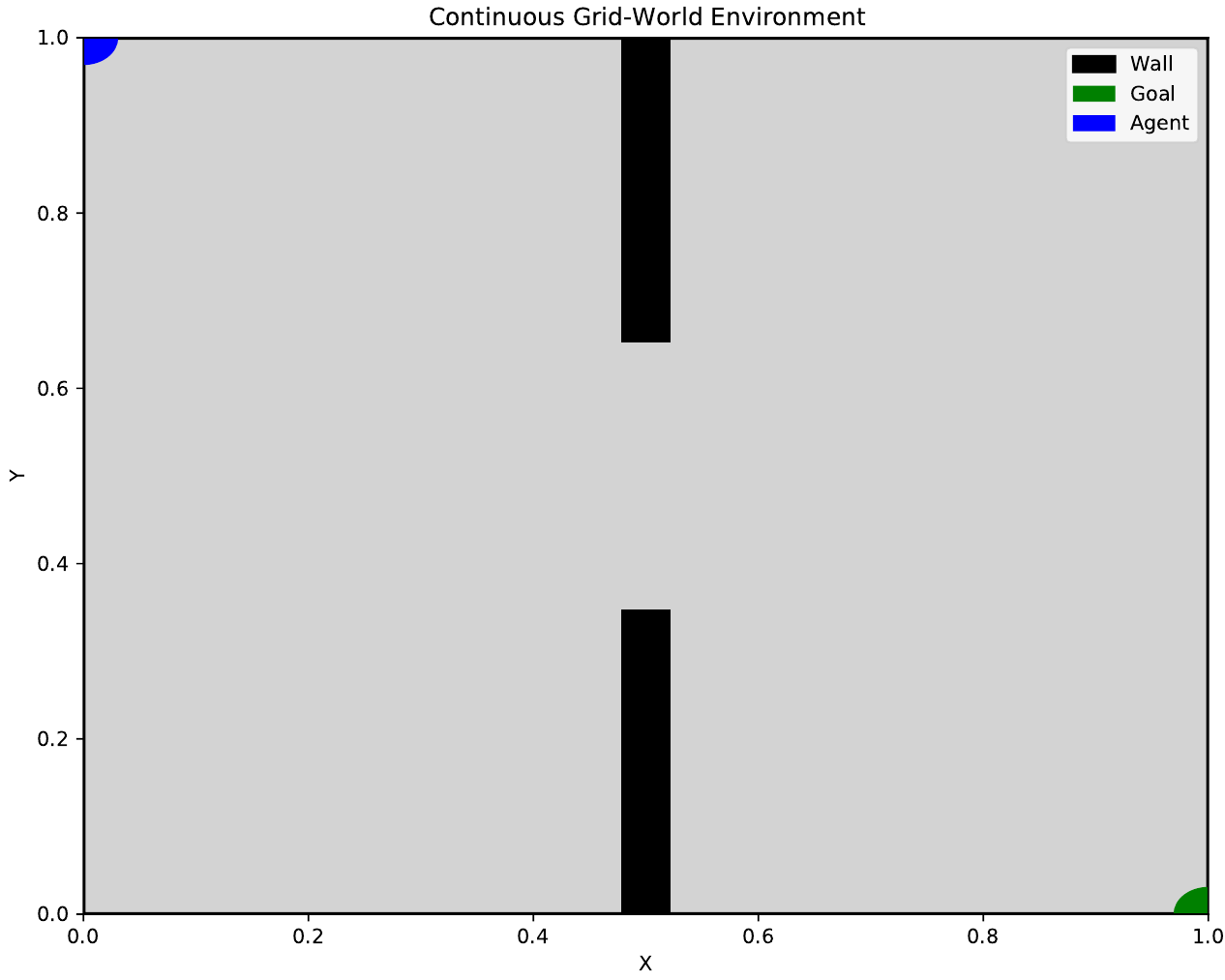} 
\caption{Environment with continuous state space used in our experiments. The goal state in this environment is specified with a green color at the right-bottom corner. The agent can start learning from different corners of the environment except for the goal state. }
\label{fig7}
\end{figure}

To test the performance of our algorithm, we designed two sets of environments with discrete and continuous state spaces. Figure \ref{fig6} and Figure \ref{fig7} indicate our gird-world environments with discrete and continuous state space, respectively. The coordinates of the agent in each step are considered as states.
The agent can do four actions navigating up, down, left, and right in both kinds of environments. 
If the agent’s action leads to collision with walls, the agent will remain in the same state. Each action can fail with a probability of $p$, in this case the agent goes to any of the neighboring states with the equal probability. We use the Euclidean distance with a maximum distance of $L=2$ for the aggregation space in all of the environments.

The agent receives a reward of $-1$ for taking each step in the environment. If the agent reaches the goal state it will get $+10$ as a reward and if it takes an action that leads to collision with walls it will get a reward of $-10$. In all of the environments, the agent can start randomly from a state at the corners of the environment, and the episode terminates if the agent reaches the goal state or if it reaches the maximum steps, defined for each environment.

As shown in Figure~\ref{fig6}, we consider 6 environments with discrete states, where the environment with a transfer state has two versions. In the first version, we consider an additional transfer action which will be fired just in the state (4, 4) and it moves the agent to the state (8, 8). This action, similar to other actions, has a probability of $p$ of failing, in which case the agent transitions to a random neighboring state. In the other version, the state (4, 4) acts like a teleport that transfers the agent to the state (8, 8). The maximum step for 2-rooms, 3-rooms, and rooms with the transfer is 100 steps, and because of the complexity of tasks in 4-rooms and 9-rooms environments, the maximum step is considered 500 in these environments. Also, the 1-room with a hallway environment is a tricky environment, so the number of maximum steps is 150 in this environment. All of the environments contain bottleneck states, because of their design which is room-to-room navigation task they have. It is clear that bottleneck states are the doorway states (in the multi-room environment) and the neighbor states of the transfer state in the environments with the transfer state.

In the environment with the continuous state space, each action results in a displacement of $0.1$ units in the corresponding coordinates of the agent's current state. For instance, the right action would move the agent from (0, 0) to (0, 0.1). After a maximum of $300$ steps, if the agent could not reach the goal state, the episode is terminated. The reward function is the same as that of the discrete environments and there is no transfer state.

\subsection{Results in Discrete State Spaces}
In our implementations, we consider the agent learns and interacts with the environment using the SARSA algorithm that has an epsilon greedy policy as its behavioral policy. The discount factor is chosen to be equal to 0.9, and the learning rate is equal to 0.99 with a decaying rate of 0.001. Also, the epsilon is 0.3 and it decays exponentially with a rate of 0.3. For parameters $\alpha$ and $\beta$ in equations \eqref{eq6}, \eqref{eq7}, and \eqref{eq8}, we choose $\alpha=4$ and $\beta=7$, that are the same parameters used in the implementation of~\cite{ghorbani2020reinforcement}. Similar to~\cite{ghorbani2020reinforcement}, we also observed that our method is not sensitive to the choice of these two parameters. All these parameters are the same for the results in all discrete-state environments and all results are an average of 10 runs. In addition, the default probability of failing an action $p$ is set equal to $33\%$ for all environments. We initialized the agent at the top-left corner of the environment in all experiments. However, our method's ability to identify bottleneck states is not dependent on the initial state, as long as the starting and goal states are in different rooms.

\begin{figure}[H]
\centering
\includegraphics[width=1\columnwidth]{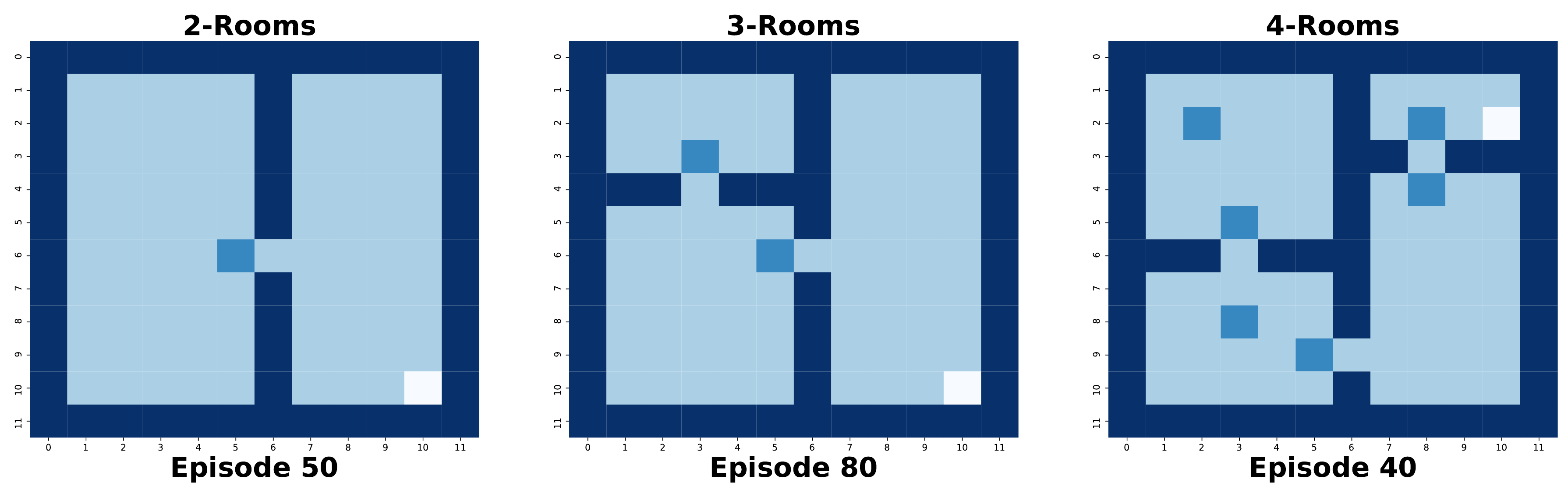} 
\caption{Performance of our proposed bottleneck detection within the high probability of action failure ($p=50\%$) in episode 50 of 2-rooms, episode 80 of 3-rooms, and episode 40 of 4-rooms environments. Identified bottlenecks (light blue states) in all of these environments are on the optimal path to reaching the goal (white state).}
\label{fig8}
\end{figure}

Figure \ref{fig3} shows the results of the model change counts and our bottleneck discovery algorithm for different number of episodes for the 2-room discrete environment. As we can see in the initial episodes we have rare model changes in different states. This is because of the exploration of the agent at the first steps. However, our agent can successfully detect the doorway bottleneck after some episodes.

Figure \ref{fig5} shows the identified subgoal/bottleneck states in different grid-world environments after 50 episodes. We can see that our agent can detect states like doorways and states around the transfer states. Also in the tricky 9-rooms environment, the agent was able to find some of the doorways that are needed to reach the final goal. In the 1-room with a hallway environment, the identified bottleneck states appear reasonable from a task decomposition perspective. They can be used to define tasks like "entering the hallway" or "leaving the hallway". However, careful tuning of the algorithm's parameters for this environment can lead to the discovery of more specific subgoals that contribute to reaching the final goal.

Because methods based on environment graphs struggle in highly stochastic environments, we tested our algorithms' ability to handle high levels of stochasticity. As shown in Figure \ref{fig8}, our algorithm successfully identified bottleneck states in environments with a high probability of action failure $p=50\%$.

\begin{figure}[H]
\centering
\includegraphics[width=1\columnwidth]{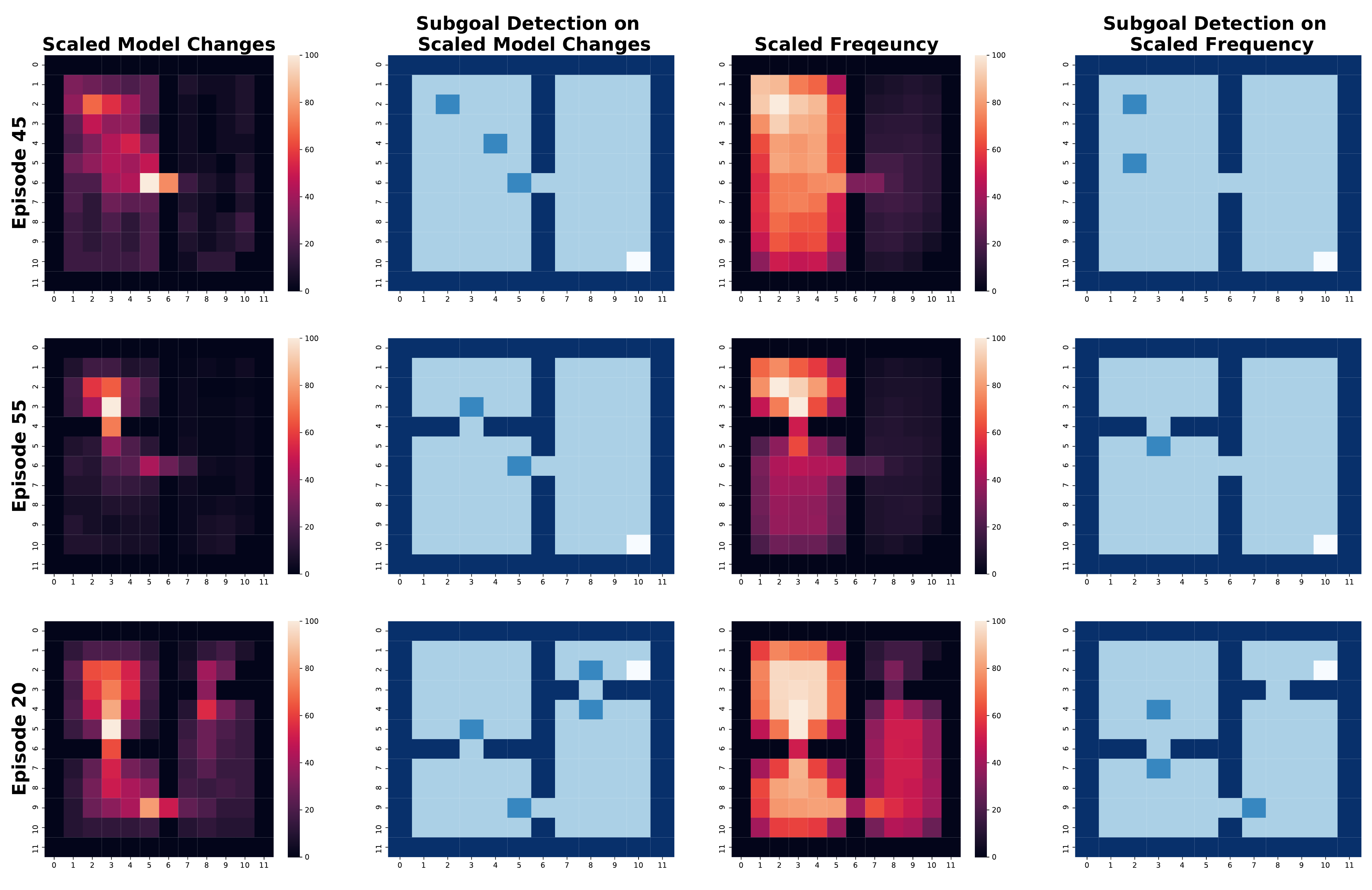} 
\caption{Performance of our proposed bottleneck detection algorithm in comparison with frequency-based algorithms in different environments. Columns one and two (from left) show the scaled model changes and the resulting bottlenecks. Columns three and four (from left) show the scaled state frequency and the detected bottlenecks. We scaled model changes and state frequencies for better comparison.}
\label{fig9}
\end{figure}

\begin{figure*}[th!]
\hspace{-0.5cm}
\includegraphics[width=1.04\textwidth]{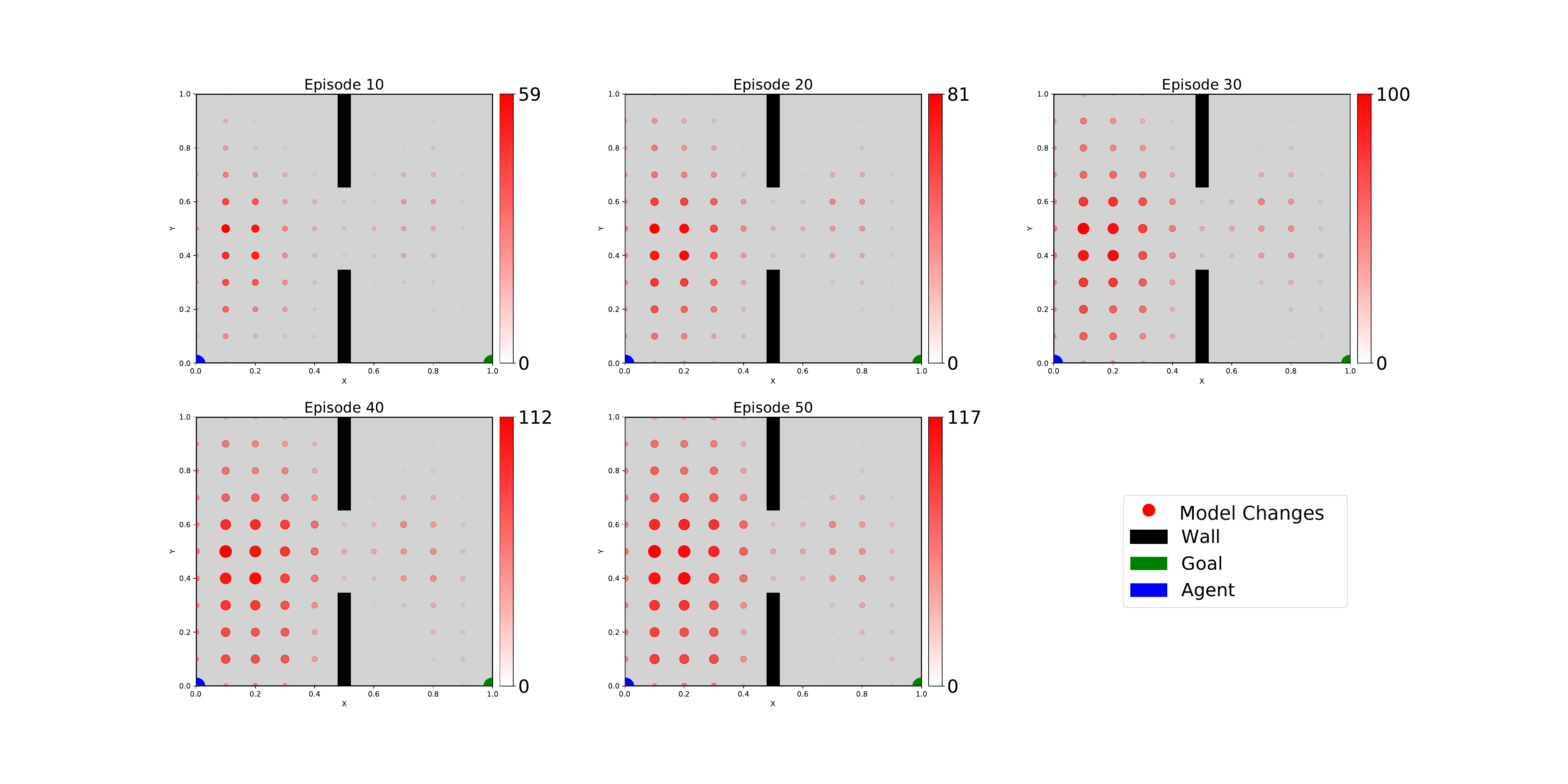} 
\caption{This figure shows the evolution of model changes in a continuous 2-room environment across episodes 10-50. The intensity of red dots represents the frequency of model changes, with brighter red indicating more changes. The agent starts from either the bottom-left or top-left corner of the left room and aims to reach the green goal in the bottom-right corner. The visualization shows an increasing concentration of model changes near the doorway connecting the two rooms, as indicated by the brighter red dots in that area.}
\label{fig10}
\end{figure*}

\subsection{Comparison with Experience-Based methods}
Since there is no quantitative criterion to compare the performance of bottleneck discovery algorithms, we show the bottlenecks discovered by our method and an experienced-based method. Similar to our method, we apply non-maximum suppression on the resulting state visit counts of the experience-based method. Figure \ref{fig9} shows the results for different environments with the action failure probability of 33\%. Because of the high level of stochasticity in these environments, relying on the trajectory information of the agent can be misleading when determining subgoal states. Therefore in such settings, the performance of experience-based methods substantially deteriorates. In contrast, our proposed method succeeds in identifying the correct subgoal states. This is because our method considers both the behavioral policy and the uncertainty in subspaces which is relatively robust to the stochasticity of the environment.

\subsection{Model Changes in Continuous State Spaces}
For the environment with continuous state space (Figure~\ref{fig7}), we used DQN ~\cite{mnih2013playing} with a fully connected architecture to learn the task. The replay buffer has the capacity of 10000  samples. The target network's weights get updated every $5$ episodes. The behavioral policy of the agent is epsilon-greedy and the parameters $\alpha$ and $\beta$ are the same as our experiments in the discrete environments.

Figure~\ref{fig10} shows the result of Algorithm~\ref{alg2:algorithm} for computing model changes in this environment. The results demonstrate that the phenomenon of model changes is not limited to the tabular settings and we have model changes when using function approximations, such as deep neural networks, for learning.

\section{Conclusion}

In this paper, we study the problem of subgoal discovery in different grid-room environments. We showed that our method can detect bottleneck states in different types of doorways and transfer states and it is robust to the noise and stochasticity of the environment. Our method does not need to save full information of trajectories in memory or to generate a graph from interactions of the agent in the environment, which can be misleading when the noise of the environment is considerably high. Our proposed method detects the bottleneck states in the environment without any supervision or predefined number.

There are several directions to expand this work in the future. One avenue is using model changes for bottleneck detection in environments with continuous states or actions, and environments with sparse rewards. Another direction of future work is searching for efficient ways to lower the time complexity of our algorithm. Making use of the discovered bottlenecks to learn reasonable and interpretable options in HRL or GCRL can be an interesting extension of the proposed method.

\bibliography{aaai24}

\end{document}